\documentclass[lettersize,journal]{IEEEtran}
\usepackage{amsmath,amsfonts}
\usepackage{algorithm}
\usepackage{algorithmic} 
\usepackage{array}
\usepackage[caption=false,font=normalsize,labelfont=sf,textfont=sf]{subfig}
\usepackage{textcomp}
\usepackage{stfloats}
\usepackage{url}
\usepackage{verbatim}
\usepackage{graphicx}
\usepackage{cite}
\usepackage{amssymb}
\usepackage{booktabs}
\usepackage{multirow}
\usepackage{multicol}
\usepackage{colortbl}

\usepackage{arydshln}
\usepackage{xcolor}
\usepackage{tabularray}
\usepackage{hyperref}
\usepackage{color,soul}

\begin{document}
\title{Federated Balanced Learning}

\author{ 
Jiaze Li$^{\ast}$,
Haoran Xu$^{\ast}$,
Wanyi Wu,
Changwei Wang,
Shuaiguang Li,
Jianzhong Ju,
Zhenbo Luo,
Jian Luan,
Youyang Qu,
Longxiang Gao,~\IEEEmembership{Senior Member,~IEEE,}
Xudong Yang,
Lumin Xing
\thanks{$^{\ast}$Equal contribution.}
\thanks{
Jiaze Li, Jianzhong Ju, Zhenbo Luo and Jian Luan are with MiLM Plus of Xiaomi Inc. Haoran Xu is with Zhejiang University.  Wanyi Wu is with Shandong University.
Changwei Wang, Shuaiguang Li, Youyang Qu and Longxiang Gao are independent researchers; Shandong Provincial Key Laboratory of Computing Power Internet and Service Computing, Shandong Fundamental Research Center for Computer Science, Jinan, 250014, China; 
Xudong Yang and Lumin Xing are with Shandong Key Laboratory of Digital Diagnosis and Treatment
of Thoracic Oncology, The First Affiliated Hospital of Shandong First Medical
University \& Shandong Provincial Qianfoshan Hospital and Shandong Engineering Research Center of Intelligent Surgery, The First Affiliated Hospital
of Shandong First Medical University \& Shandong Provincial Qianfoshan
Hospital, Jinan, China.

}
\thanks{This work was supported by the Shandong Provincial Key R\&D Program No.2025KJHZ013,  National Key R\&D Program of China No.2025YFE0216800, Shandong Provincial University Youth Innovation and Technology Support Program No.2022KJ291, Shandong Provincial Natural Science Foundation for Young Scholars Program No.ZR2025QC1627 and Qilu University of Technology (Shandong Academy of Sciences) Youth Outstanding Talent Program No. 2024QZJH02, Shandong Provincial Natural Science Foundation for Young Scholars (Category C) No.ZR2025QC1627.
}
}

\markboth{IEEE Transactions on Multimedia,~Vol.~XX, No.~XX, March~2026}%
{Shell \MakeLowercase{\textit{et al.}}: A Sample Article Using IEEEtran.cls for IEEE Journals}

\maketitle

\begin{abstract}
Federated learning is a paradigm of joint learning in which clients collaborate by sharing model parameters instead of data. However, in the non-iid setting, the global model experiences client drift, which can seriously affect the final performance of the model. Previous methods tend to correct the global model that has already deviated based on the loss function or gradient, overlooking the impact of the client samples. In this paper, we rethink the role of the client side and propose Federated Balanced Learning, i.e., \textbf{FBL}, to prevent this issue from the beginning through sample balance on the client side. Technically, \textbf{FBL} allows unbalanced data on the client side to achieve sample balance through knowledge filling and knowledge sampling using edge-side generation models, under the limitation of a fixed number of data samples on clients. Furthermore, we design a Knowledge Alignment Strategy to bridge the gap between synthetic and real data, and a Knowledge Drop Strategy to regularize our method. Meanwhile, we scale our method to real and complex scenarios, allowing different clients to adopt various methods, and extend our framework to further improve performance. Numerous experiments show that our method outperforms state-of-the-art baselines. The code is released upon acceptance.
\end{abstract}


\section{Introduction}
\label{sec:intro}

Federated learning (FL)~\cite{DBLP:conf/aistats/McMahanMRHA17} allows distributed clients~\cite{bagci2017compete} to collaboratively train a shared global model using their local \textit{multimedia data}—such as images, videos, and audio—without uploading raw data to a centralized server. This paradigm not only preserves user privacy but also significantly reduces communication overhead, which is especially critical when handling large-volume multimedia content.

However, current FL systems suffer from severe performance degradation due to the non-independent and identically distributed (non-IID) nature of client-held multimedia data. For instance, user-generated images or videos often exhibit strong domain bias (e.g., specific lighting conditions, object categories, or recording environments), leading to skewed local distributions. Such heterogeneity can cause the global model to overfit to dominant client modalities or domains, resulting in suboptimal generalization across diverse multimedia scenarios. Numerous approaches have been proposed to mitigate this issue, including FedProx~\cite{DBLP:conf/mlsys/LiSZSTS20}, SCAFFOLD~\cite{pmlr-v119-karimireddy20a}, and MOON~\cite{li2021model}. Nevertheless, these methods primarily operate at the optimization level—adjusting gradients or loss functions to correct model drift after it occurs—rather than addressing the root cause: the imbalance and scarcity of representative multimedia samples across clients. Moreover, under the traditional FL assumption of resource-constrained edge devices, most existing works overlook the potential for on-device multimedia model enhancement.

Recent breakthroughs in generative models, particularly diffusion models~\cite{ho2020denoisingdiffusionprobabilisticmodels,song2022denoisingdiffusionimplicitmodels,oord2018neuraldiscreterepresentationlearning}, have enabled the synthesis of high-fidelity \textit{multimedia content}, especially photorealistic images. Models like Stable Diffusion~\cite{rombach2022highresolutionimagesynthesislatent} generate visual data that is substantially more realistic and semantically coherent than outputs from earlier generative frameworks such as GANs~\cite{goodfellow2014generativeadversarialnetworks}, VAEs~\cite{Kingma_2019}, or transformer-based synthesizers~\cite{esser2021tamingtransformershighresolutionimage}. This advancement makes synthetic multimedia data a viable surrogate for real user data in privacy-sensitive learning scenarios.

Concurrently, the computational capabilities of edge devices are rapidly evolving, enabling the deployment of sophisticated generative models directly on multimedia-capable hardware. For example, SnapFusion~\cite{yu2025comp} and MobileDiffusion~\cite{zhao2024mobilediffusioninstanttexttoimagegeneration} have successfully quantized and optimized Stable Diffusion for mobile platforms, allowing second-scale generation of high-quality images that rival those produced by cloud-based counterparts~\cite{liu2025potential}. Even in highly constrained settings, recent work has demonstrated Stable Diffusion inference on a Raspberry Pi with only 260\,MB of RAM. These developments open a new avenue for FL: instead of merely consuming data, edge clients can now \textit{generate} diverse, synthetic multimedia samples to balance their local distributions.

This leads us to rethink the role of the client in federated multimedia learning—not just as a passive data holder, but as an active participant capable of enriching its local dataset through on-device generative modeling. By leveraging such capabilities, we aim to proactively mitigate the non-IID problem at the \textit{data level}, specifically from the perspective of multimedia sample balance~\cite{zhao2025constructing}.



In this paper, we propose Federated Balanced Learning (FBL), with employing edge-side generative model for knowledge filling to achieve sample balance, delve into the details and ask two questions:

\textit{Q1: How to achieve sample balance on client side?}
In a naive approach, we can generate samples using edge-side generative models to complete the number of all classes of samples to the maximum number of classes to achieve sample balance. The disadvantage of the above method is that in theory there is no upper limit to the amount of synthetic data it needs to generate. Moreover, in real scenarios, the usage duration of computational resources on the client side is limited. To improve the generality, we propose the knowledge filling strategy under constrained computation. In constrained computation, the number of client samples does not increase and in this scenario, the duration of available computing resources on the client side also remains unchanged. Therefore, our solution firstly divides the classes on every client into data-excessive, data-scarce, and data-missing classes based on the expected balanced number of samples. Then, knowledge sampling is applied to data-excessive classes, and knowledge filling is used for data-scarce and data-missing classes to maintain sample balance under constrained computation.

\textit{Q2: How can the gap between synthetic data and real data be bridged?} Considering the domain gap~\cite{yu2023narrowing} between synthetic data generated by the model and real data, we propose a Knowledge Alignment Strategy. We alse design a Knowledge Drop Strategy to regularize the method.

Furthermore, we scale our method \textbf{FBL} into real complex scenarios where the duration of computational resources among clients may be unbalanced and extend \textbf{FBL} our framework to further improve performance. Finally, we conducted experiments with real-world datasets under other domains, such as satellite datasets~\cite{ zeng2022geo}, and discussed the existing limitations and future prospects of \textbf{FBL}.



We summarize the main contributions as follows:
\begin{itemize}
\item We rethink the role of the client side and propose \textbf{FBL} to prevent the non-iid problem from the beginning through sample balance on the client side.

\item We design a Knowledge Alignment Strategy to bridge the gap between synthetic and real data, and a Knowledge Drop Strategy to regularize our method. Extensive experiments validate that our approach achieves best results.

\item We extend our framework to further improve performance in real and complex scenarios.
\end{itemize}

\section{Related Works}
\label{sec:related}

Previous research has addressed the challenge of non-IID data in federated learning. FedProx~\cite{ma2023fedsh} incorporates an additional term into the local model's objective function. Works such as SCAFFOLD~\cite{pmlr-v119-karimireddy20a} address client drift by employing control variates for gradients, albeit without directly considering inconsistencies between local and global objectives. MOON~\cite{li2021model} utilizes the similarity between model representations to correct the local training of individual parties. FedGen~\cite{zhu2021datafreeknowledgedistillationheterogeneous} enables the server to generate features that conform to the global data distribution by sharing the labels of clients and distilling globally. However, the sharing of client labels poses privacy risks~\cite{ xu2023federated}.

\noindent\textbf{Generative Models in FL.}
With the development of generative models~\cite{ramesh2022hierarchicaltextconditionalimagegeneration,saharia2022photorealistic,nichol2022glidephotorealisticimagegeneration}, there is a renewed awareness of the importance of synthetic data and generative models. Generating high-quality images based on text prompts has become a new paradigm. Therefore, a few works have explored the usage of synthetic data, especially images as training data. For example, someone~\cite{wang2024generateddatahelpcontrastive} explore the ratio between synthetic data and real data. On the other hand, synthetic data and generative models are not sufficiently explored in the FL setting. FGL~\cite{liu2025context} trains the model entirely using synthetic data~\cite{ websdale2021speaker}, with the disadvantage of using global tag sharing, and having privacy risk~\cite{ xu2023federated}. 

Therefore, how to integrate the existing edge-side generative model in FL without privacy concerns is a key challenge. \textit{Our solution is for each client to achieve sample balance to prevent the non-IID issue from the beginning by using edge-side generative models.}

\section{Preliminary}\label{sec:preliminary}
In federated learning (FL), a system with \( K \) clients is considered, where each client possesses its own dataset \( \mathcal{D}_k = \{(x_i, y_i)\}_{i=1}^{N_k} \). Here, \( x_i \) represents a data sample and \( y_i \) represents the corresponding label. The goal is to collaboratively train a global model \( \theta \) that minimizes the aggregate loss across all clients' data without requiring direct access to each client's data.

The objective function for traditional federated learning methods is formulated as:
\begin{equation}
\min_{\theta} \frac{1}{K} \sum_{k=1}^{K} \mathcal{L}_{\mathcal{D}_k}(\theta),
\end{equation}
where \( \mathcal{L}_{\mathcal{D}_k}(\theta) \) is the expected loss over the data distribution \( \mathcal{D}_k \) of the \( k \)-th client:
\begin{equation}
\mathcal{L}_{\mathcal{D}_k}(\theta) = \mathbb{E}_{(x_i, y_i) \sim \mathcal{D}_k}[\ell(h(x_i; \theta), y_i)],
\end{equation}
with \( \ell \) being a cross-entropy loss function and \( h(x_i; \theta) \) the model's prediction for input \( x_i \).


\begin{figure*}[t]
\centering
\includegraphics[width=0.8\linewidth]{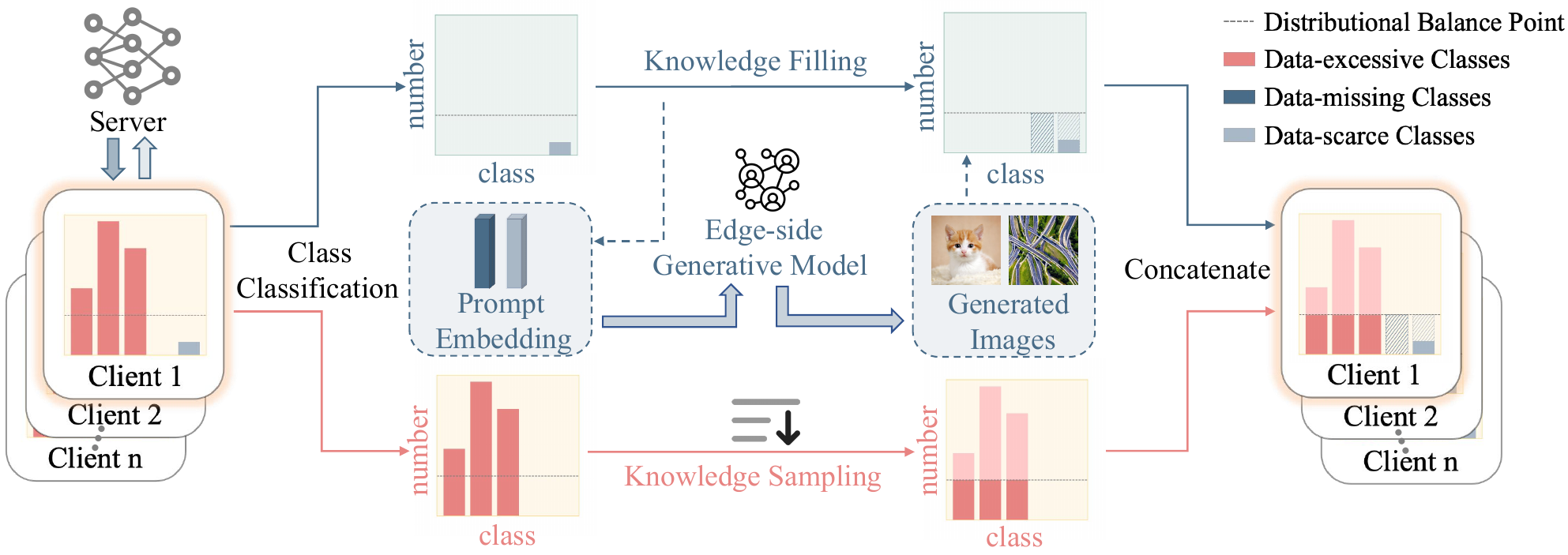}
\caption{Illustration of \textbf{FBL}. Our framework mainly includes two parts: Knowledge Filling and Knowledge Sampling. Before these two parts, our framework first needs to determine the distributional balance point and then classify all classes into three categories: data-excessive, data-scarce, and data-missing. For data-excessive classes, we use knowledge sampling, while for data-scarce and data-missing classes, we utilize knowledge Filling.}
\label{fig:pipeline}
\end{figure*}

\section{Methodology}\label{sec:method}
In this section, we present our approach in four parts: Class Classification, Knowledge Filling, Knowledge Sampling, and Extended and Unified Framework.

Specifically, in our algorithm, each client first divides all classes into three categories: data-excessive, data-scarce, and data-missing. We then apply knowledge sampling for the data-excessive classes and knowledge filling for the other two categories. Finally, we further expand our framework in the Extended and Unified Framework section. Our framework is shown in Figure \ref{fig:pipeline}, and the detailed workflow is presented in Algorithm~\ref{alg:1}.

\subsection{Class Classification} \label{category_classification}
In constrained computation settings, achieving sample balance while ensuring that the number of samples per client does not exceed the original amount is crucial. To address this, we designate the mean of the sample counts across all classes in each client as the distributional balance point. Mathematically, this balance point is calculated as follows:
\begin{equation}\label{classification}
B_k = \left\lfloor \frac{1}{C} \sum_{c=1}^{C} n_k^c \right\rfloor 
\end{equation}
where \( B_k \) represents the distributional balance point of the \( k \)-th client, \( C \) denotes the total number of classes, and \( n_k^c \) indicates the number of samples in class \( c \). The operator \( \left\lfloor \cdot \right\rfloor \) denotes the floor function, which ensures that the calculated mean is rounded down to the nearest integer. This method ensures that the distribution of samples is balanced across all clients without increasing the overall sample size.

To further categorize the data based on this balance point, classes with sample counts exceeding the balance point are termed data-excessive classes. Conversely, classes with sample counts less than the balance point but greater than zero are defined as data-scarce classes. Classes with no samples are classified as data-missing classes.

\subsection{Knowledge Filling} \label{data_filling}
\begin{figure}[h]
    \centering 
    \includegraphics[width=1\linewidth]{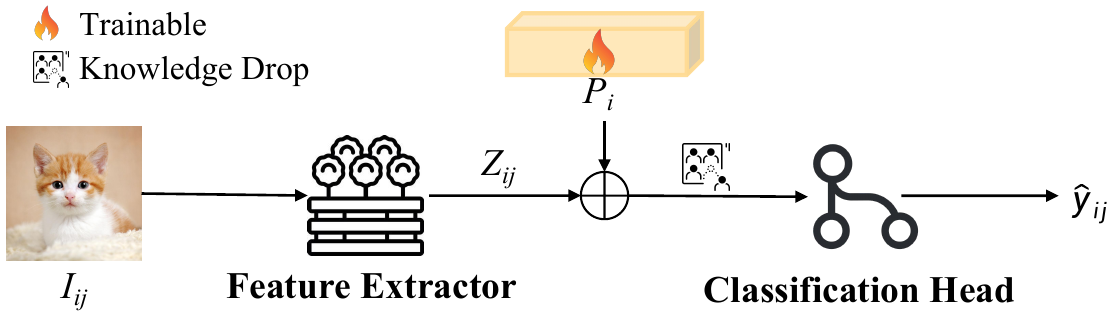}
    \caption{Illustration of the Knowledge Alignment Strategy and Knowledge Drop Strategy.}
    \label{images:embedding}
\end{figure}

To achieve sample balance, particularly for data-scarce and data-missing classes, we employ the realistic generation capabilities of edge-side Stable Diffusion, with specific prompts detailed in the Appendix, to augment the sample numbers of these classes to the distributional balance point. To address the domain gap between generated images and real-world client images, we propose two strategies: Knowledge Alignment Strategy and Knowledge Drop Strategy. The former aims to narrow the domain gap between synthetic data and real data, while the latter introduces a regularization technique similar to Dropout~\cite{zhong2021dynamic} to prevent overfitting in the Knowledge Alignment Strategy, as shown in Figure~\ref{images:embedding}.

\noindent\textbf{Knowledge Alignment Strategy.}
The domain gap between the synthetic data and the real data may adversely affects model performance, rendering it suboptimal. To address this issue, we propose a Knowledge Alignment strategy. Specifically, for each class \( i \) within the categories of data-scarce and data-missing classes, we create a learnable embedding \( P_i \) to bridge the gap between the generated domain and the real domain. Following the popular settings in previous federated learning works, we use image classification tasks as an example, where a typical image classification network, such as ResNet-18~\cite{he_deep_2016}, usually comprises a feature extractor \( F \) and a classification head \( H \). An image, when passed through the feature extractor \( F \), yields a prototype, which, after being processed by the classification head, results in the final prediction. Inspired by FedDBE~\cite{zhang2023eliminating}, we augment the generated image prototypes with learnable embeddings and then process them through the classification head. The formulation is as follows:

\begin{equation}
\hat{y}_{ij} = H(F(I_{ij}) \oplus P_i)
\end{equation}
where \( P_i \) is the learnable embedding for class \( i \), and \( I_{ij} \) is the \( j \)-th generated image of class \( i \). The image \( I_{ij} \) is first passed through the feature extractor \( F \) to obtain a prototype, which is then element-wise added to the embedding \( P_i \). The resultant vector is subsequently fed into the classification head \( H \) to produce the model's prediction \( \hat{y}_{ij} \). This embedding-based augmentation method effectively compensates for the domain gap, thereby enhancing the overall model performance.

\noindent\textbf{Knowledge Drop Strategy.}
To further improve the robustness of the model and prevent overfitting during the learning of embeddings, we introduce a Knowledge Drop Strategy. This strategy involves randomly dropping a subset of samples within each batch, thereby excluding their embeddings during training. Let \( \zeta \) be the drop rate, representing the proportion of samples to be dropped. For a given batch size \( B \), the number of samples to be dropped is \( \zeta B \). The formulation for the Knowledge Drop Strategy is as follows:

\begin{equation}
\hat{y}_{ij} = 
\begin{cases} 
H(F(I_{ij}) \oplus P_i) & \text{if } I_{ij} \text{ is not dropped} \\
H(F(I_{ij})) & \text{if } I_{ij} \text{ is dropped}
\end{cases}
\end{equation}

where \( I_{ij} \) is the \( j \)-th image of class \( i \) in the batch. If \( I_{ij} \) is selected to be dropped, its embedding \( P_i \) is not added to the prototype. This random dropping of embeddings, controlled by the drop rate \( \zeta \), helps in regularizing the model and mitigating overfitting, as shown in Figure~\ref{images:embedding}.

\subsection{Knowledge Sampling}
\label{importance_sampling}
In scenarios characterized by data-excessive classes, it is imperative to employ sampling methods to downsample the number of samples to a distributional balance point. Simultaneously, we must ensure that the performance of the model on each client remains as close as possible to the performance observed prior to sampling. The target is to minimize the performance discrepancy between the models trained on the pre-sampled and post-sampled data, ensuring that the diversity and representation of data are preserved as closely as possible. To achieve this, we optimize the following objective:
\begin{equation}
\begin{aligned}
 \min_{\mathcal{M}} \bigg| &f_p\left(\mathcal{M}(\sum_{i=1}^{C_e} \sum_{j=1}^{S_i}X_{ij}), \sum_{l=1}^{C_s} \sum_{m=1}^{S_{l}^{'}} X_{lm} \right) \\
 &- f_p\left( \sum_{i=1}^{C_e} \sum_{j=1}^{S_i} X_{ij}, \sum_{l=1}^{C_s} \sum_{m=1}^{S_{l}^{'}} X_{lm} \right) \bigg| 
\end{aligned}
\end{equation}
where \( \mathcal{M} \) defines the sampling operation applied to the data-excessive class collection. \( S_i \) represents the set of samples from the \( i \)-th data-excessive classes and \( C_e \) defines the total  number of data-excessive classes with \( X_{ij} \) denoting the \( j \)-th sample in the \( i \)-th data-excessive class. 
On the other side, \( C_s \) defines the total  number of special classes.
\( S_{l}^{'} \) denotes the set of samples from the special classes, which include data-scarce and data-missing classes, with \( X_{lm} \) denoting the \( m \)-th image in the \( l \)-th special class. 

Here, \( f_p(\cdot, \cdot) \) represents the model's performance function, which takes both the aggregated \( X_{ij} \) and \( X_{lm} \) as inputs, evaluating the combined effect on model performance. This approach ensures that the model maintains robust performance across all clients, even after the sampling process.


\noindent\textbf{Knowledge Sampling Based on Loss.}
The naive sampling method for data-excessive classes is random sampling, indiscriminately selecting from the available class samples under the assumption that each sample contributes equally to model training. However, previous work ~\cite{Goyal_2024_CVPR,yoo2019learninglossactivelearning,abbas2023semdedupdataefficientlearningwebscale} demonstrates that the importance of different samples to model training can vary significantly. This variability can be measured using the model's loss for each sample.

Inspired by this insight, we propose a loss-based knowledge sampling technique. Our method prioritizes samples that result in higher losses, asserting that these are more crucial for effective training. The selection process is formalized as follows:
\begin{equation}
\max \left(\sum_{j=1}^{B_k} \mathcal{L}(X_{ij})\right), X_{ij} \in S_i, S_i \subseteq \{S_i\}_{i=1}^{C_e}
\end{equation}
Here, \( C_e \) indicates the total number of data-excessive classes, \( S_i \) is the set of samples for the \( i \)-th data-excessive class, and \( B_k \) represents the number of balance point for the \( k \)-th client. \( \mathcal{L}(\cdot) \) defines the loss function, while \( X_{ij} \) is the \( j \)-th sample from \( i \)-th data-excessive class. The formula characterizes that all data-excessive classes in the \( k \)-th client have to pick the top \( B_k \) samples by loss function \( \mathcal{L}(\cdot) \).

\noindent\textbf{Knowledge Replay.} 
Previous research ~\cite{Goyal_2024_CVPR,hoffmann2022trainingcomputeoptimallargelanguage} has demonstrated that the importance of the same sample varies at different stages of the model's training. Inspired by this insight, and to mitigate the loss of knowledge diversity caused by downsampling, we employ a strategy of periodic knowledge replay every \( m \) rounds. Meanwhile, continuous learning research ~\cite{rolnick2019experiencereplaycontinuallearning,li2024efficientreplayfederatedincremental} indicates that when a model is trained on a new dataset, it tends to catastrophically forget the old data. To address the forget issue, we also propose a knowledge replay strategy. We define the precondition operation as follows:
\begin{equation}
 g(\mathcal{R}_{i}^{t}, q) := \{x_1, x_2, \ldots, x_q\}, x_q \in \mathcal{R}_{i}^{t}
\end{equation}
 The operation \( g(\mathcal{R}_{i}^{t}, q) \) defines a set of data after sampling \( q \) samples from  \(\mathcal{R}_{i}^{t} \), then,we formulate the knowledge replay strategy:
\begin{equation}
 \mathcal{R}_{i}^{t+1} = g(\mathcal{R}_{i}^{t},\gamma B_k) \cup g((S_i \setminus \mathcal{R}_{i}^{t}),(1 - \gamma)B_k) 
\end{equation}
Here, \( \mathcal{R}_{i}^{t} \) represents the sampled dataset in  \( t \)-th cycle the for the \( i \)-th data-excessive class. \( \gamma\) is a parameter ranging from 0 to 1 that indicates the proportion of samples retained from the previous cycle's dataset and \( S_i \) denotes the set of samples for the \( i \)-th data-excessive class. \( S_i \setminus \mathcal{R}_{i}^{t} \) denotes the set of new samples are drawn from the remaining dataset which the set of elements after \( S_i \) removes the elements from \( \mathcal{R}_{i}^{t} \). The knowledge replay strategy judiciously combines a portion of previously selected samples with new samples.

\begin{algorithm}[tb]
    \caption{\textbf{FBL} Algorithm}
    \label{alg:1}
    \renewcommand{\algorithmicrequire}{\textbf{Input:}}
    \renewcommand{\algorithmicensure}{\textbf{Output:}}
	\begin{algorithmic}[1] 
            \REQUIRE  global rounds $T$, clients number $K$, sample number $n_k^c$  in class $c$, local epoch $E$, learning rate $\eta$
		\ENSURE Trained global model $\bf{w}$ 
		\STATE {initialize and distribute $\bf{w}_{0}$ for client $k$, $k{\in}[1, K]$}
        \STATE {initialize $B_k = \left\lfloor \frac{1}{C} \sum_{c=1}^{C} n_k^c \right\rfloor $}
        \STATE {obtain data-excessive class number $C_e$, special class number $C_s$  based on $B_k$ and $n_k^c$}
		\FOR {$t=1$ to $T$}
		\FOR {$k=1$ to $K$ in parallel}
            \FOR {$i=1$ to $C_e$}
            \STATE {Execute knowledge sampling to make number of sample $S_i$ down to $B_k$}
            \ENDFOR
            \FOR {$j=1$ to $C_s$}
            \STATE {Execute knowledge filling to make number of sample $S_j^{'}$ up to $B_k$}
            \ENDFOR
            \FOR {local epoch $1$ to $E$}
            \STATE {$\mathbf{w}^k = \mathbf{w}^k - \eta \nabla_{\mathbf{w}^k}f_n(\mathbf{w}^k)$}
            \ENDFOR
		\ENDFOR
        \STATE {Aggregate local models: $\mathbf{w}_{t+1} = \frac{1}{K}\sum_{k=1}^K\mathbf{w}^k$}
		\ENDFOR
		\STATE \textbf{return} global model $\bf{w}_{t}$
	\end{algorithmic}
\end{algorithm}
\subsection{Extended and Unified Framework}
\label{unified_framework}
\begin{figure}[h]
    \centering 
    \includegraphics[width=0.8\linewidth]{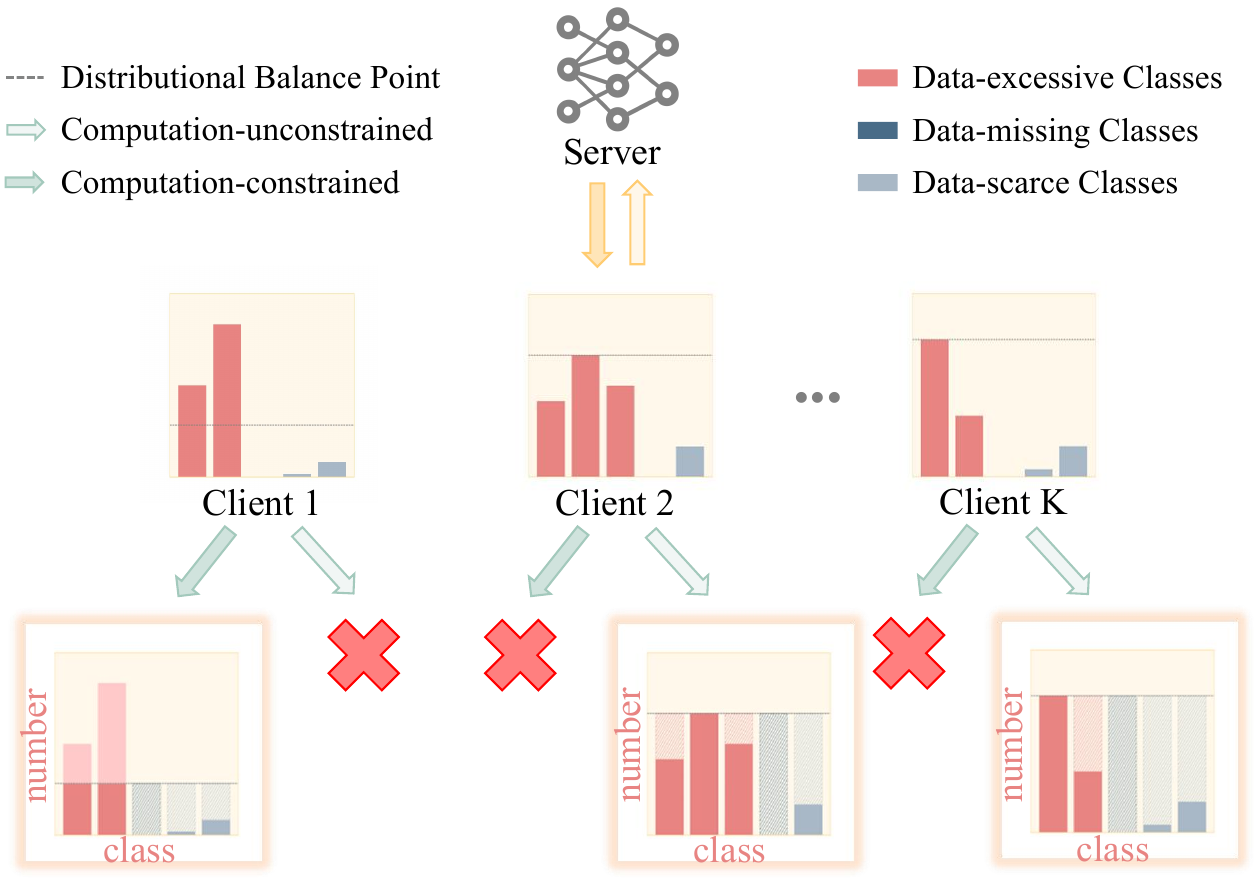}
    
    \caption{Extended and Unified Framework. Clients can self-decide to use either computation-constrained or computation-unconstrained resources based on the duration of available computing resources.}
    \label{images:framework}
\end{figure}

In considering real-world or more complex scenarios where the duration of available computing resources on the client is unevenly distributed, with some clients having no such limitations on the duration of available computing resources, we extend and propose a computation-aware client self-decision framework. As depicted in Figure~\ref{images:framework}, under this framework, clients autonomously decide on knowledge filling strategies based on their computational resources, distinguishing between computation-constrained and computation-unconstrained scenarios. In computation-constrained situations, as shown in Eq.~\ref{classification}, the distributional balance point is the mean of samples across all classes. Conversely, in computation-unconstrained situations, the balance point is the maximum value among samples of all classes, which can be defined as follows:
\begin{equation}
B_k = \max_{c \in C} (n_k^c)
\end{equation}
where \( n_k^c \) denotes the number of samples in the \( c \)-th class, and \( C \) denotes the total number of classes.

We define \(\lambda\) as the proportion of computation-unconstrained clients, where \(\lambda\) ranges between 0 and 1. To validate the efficacy of our framework, we conducted experiments under varying \(\lambda\) values. Details of experiments are shown later in Section~\ref{unfied}.


\subsection{Unified Perspective and Generalization of FBL}  
We propose that FBL follows the "Sampling-Generation-Alignment" framework. Specifically, we use a loss-based method to perform knowledge sampling for data-excessive classes. For data-missing and data-scarce classes, we employ an edge-side generation model to synthesize knowledge. Subsequently, the Knowledge Alignment Strategy and Knowledge Drop Strategy are applied to align the generated knowledge with real-world knowledge. Our methodology is evaluated within the popular federated learning paradigm using image classification tasks. However, the FBL approach based on the "Sampling-Generation-Alignment" framework is not limited to this domain. For example, in natural language processing (NLP) tasks, we can leverage large language models (LLMs) to generate knowledge, while also integrating other components of our framework to enhance task performance.

\section{Experiments}
\label{sec:exp}
\subsection{Experiment Setting} 
\label{subsec:experiment_settings}
\noindent\textbf{Dataset.} 
To assess the performance of our proposed federated learning methodology, we conducted experiments on three distinct datasets: CIFAR-10~\cite{krizhevsky2009learning}, CIFAR-100~\cite{krizhevsky2009learning}, and Tiny ImageNet~\cite{chrabaszcz2017downsampledvariantimagenetalternative}. To model the heterogeneous data distribution typical of real-world scenarios across multiple clients, we utilized the Dirichlet distribution technique~\cite{lin2020ensemble}. The heterogeneity level was controlled by setting the Dirichlet parameter $\alpha$ to \{0.1, 0.3, 0.5\}. Lower $\alpha$ values simulate increased distributional heterogeneity among clients, providing a stringent test under challenging conditions.

\noindent\textbf{Model Architecture.} 
In our experiments, we utilize ResNet-18 as the backbone for all datasets. ResNet-18 consists of 18 layers, including convolutional layers, batch normalization layers, ReLU activation functions, and pooling layers. It features residual blocks that incorporate skip connections, allowing the network to maintain performance even with increased depth.

\noindent\textbf{Baselines.} 
Beside of FedAvg~\cite{DBLP:conf/aistats/McMahanMRHA17}, we also compare against various types of non-iid federated learning methods with the proposed approach in our experiments, such as FedProx~\cite{DBLP:conf/mlsys/LiSZSTS20}, Scaffold~\cite{pmlr-v119-karimireddy20a}, MOON~\cite{li2021model}, and FedDyn~\cite{acar2021federatedlearningbaseddynamic}. In addition, the knowledge-distillation-based method, FedGen~\cite{pmlr-v139-zhu21b} and FedNTD~\cite{NEURIPS2022_fadec8f2} are also as our baselines. As a method of cross-rounds aggregation,
FedCDA~\cite{wang2024fedcda} is our baseline too.

\noindent\textbf{Implementation Details.}
All experiments were conducted on RTX 3090 GPUs. We utilized the 4-bit quantized version of Stable Diffusion 2-1, which can run on mobile devices. We performed the experiments using 20 clients over 200 rounds. In each round, 50\% of the clients were randomly selected for 10 iterations of local training. We configured the parameter \(m\) to 50, set \(\gamma\) to 0.1, used a batch size of 64, and set \(\zeta B = 2\). The optimization utilized the SGD optimizer with a learning rate of 1e-3, momentum of 1e-4, and weight decay of 1e-5.

\subsection{Performance Evaluation}
We demonstrate the mean accuracy of each baseline and the proposed method \textbf{FBL} under various heterogeneity levels in Table~\ref{tab:result1}.

\begin{table*}[ht]
\caption{Mean accuracy in the scenario over three datasets on various heterogeneity levels. \textbf{Bold} fonts highlight the best accuracy. Underline highlights the second best accuracy. Ours(25\%) refers to our method extended to scenarios where 25\% of the clients are computation-unconstrained, while Ours(50\%) refers to our method extended to scenarios where 50\% of the clients are computation-unconstrained.}
\label{tab:result1}
\centering
\footnotesize
\setlength{\tabcolsep}{4pt} 
\begin{tabular}{lccccccccc}
\toprule
 & \multicolumn{3}{c}{CIFAR-10} & \multicolumn{3}{c}{CIFAR-100} & \multicolumn{3}{c}{Tiny-ImageNet} \\
\cmidrule(r){2-4} \cmidrule(r){5-7} \cmidrule(r){8-10}
$\alpha$ & 0.1 & 0.3 & 0.5 & 0.1 & 0.3 & 0.5 & 0.1 & 0.3 & 0.5 \\
\midrule
FedAvg~\cite{DBLP:conf/aistats/McMahanMRHA17} & 56.60(1.56) & 65.96(0.85) & 69.25(0.52) & 33.89(0.86) & 37.68(0.67) & 39.86(0.75) & 28.04(0.35) & 31.37(0.13) & 32.69(0.23) \\ 
FedProx~\cite{DBLP:conf/mlsys/LiSZSTS20} & 52.46(2.08) & 71.88(2.87) & 78.67(1.89) & 30.89(2.16) & 40.68(1.28) & 48.88(1.89) & 22.88(1.19) & 30.78(0.23) & 34.72(1.06) \\ 
Scaffold~\cite{pmlr-v119-karimireddy20a} & 59.81(2.28) & 77.70(2.51) & 79.57(2.02) & 35.88(0.72) & 45.79(0.28) & 49.36(0.76) & 28.37(0.19) & 34.02(0.89) & 36.88(1.02) \\ 
MOON~\cite{li2021model} & 51.38(5.21) & 74.87(2.96) & 79.42(0.57) & 36.18(0.42) & 45.07(0.44) & 50.57(0.54) & 26.78(0.86) & 32.78(1.22) & 36.44(1.32) \\ 
FedGen~\cite{pmlr-v139-zhu21b} & 42.14(1.21) & 50.11(1.32) & 59.17(0.88) & 27.66(1.16) & 33.34(1.34) & 37.78(1.04) & 25.66(0.19) & 28.64(0.46) & 31.44(1.25) \\ 
FedDyn~\cite{acar2021federatedlearningbaseddynamic} & 56.75(0.10) & 78.85(1.35) & 82.95(0.17) & 34.97(0.84) & 46.10(0.13) & 50.89(0.34) & 27.98(0.39) & 34.45(0.56) & 37.78(0.98) \\ 
FedNTD~\cite{NEURIPS2022_fadec8f2} & 54.75(1.12) & 78.66(0.98) & 80.48(0.48) & 34.48(0.88) & 45.77(1.06) & 51.16(1.38) & 28.22(0.16) & 34.08(0.66) & 37.58(0.44) \\ 
FedCDA~\cite{wang2024fedcda} & 59.98(0.22) & 76.78(0.28) & 78.88(0.14) & 32.48(0.16) & 44.76(0.24) & 50.45(0.22) & 26.44(0.13) & 33.11(0.14) & 35.88(0.17) \\ 
\midrule
Ours & 76.06(0.24) & 81.62(0.30) & 84.58(0.08) & 36.52(0.30) & 48.65(0.20) & 52.29(0.16) & 25.61(0.20) & 35.60(0.16) & 40.32(0.22) \\
Ours(25\%) & \underline{79.21}(0.15) & \textbf{83.88}(0.14) & \underline{84.95}(0.28) & \underline{48.19}(0.37) & \textbf{53.52}(0.29) & \textbf{57.61}(0.45) & \underline{34.68}(0.38) & \underline{40.18}(0.26) & \textbf{46.58}(0.34) \\
Ours(50\%) & \textbf{79.88}(0.19) & \underline{83.28}(0.21) & \textbf{85.89}(0.26) & \textbf{48.47}(0.26) & \underline{53.50}(0.42) & \underline{57.36}(0.56) & \textbf{35.72}(0.29) & \textbf{40.39}(0.20) & \underline{46.34}(0.55) \\
\bottomrule
\end{tabular}
\end{table*}

Table~\ref{tab:result1} shows the performance of our method and baselines. We can find that our approach has achieved optimal results in almost all non-iid settings. Meanwhile, if we release the constrained computation, we find that 
when the proportion of resource-unconstrained clients \(\lambda\) is 50\% or 25\% , Ours(50\%) and Ours(25\%) cover best and second best results in all settings of three datasets. In particular, on the setting of CIFAR-10(Dir 0.1), the performance of our method(Ours(50\%)) outperforms the baseline method of best performance by nearly 33\%. This may be that CIFAR-10 dataset has a relatively small number of classes and the knowledge alignment embedding is easier to converge under 200 global rounds. At the same time, the performance of FedGen method is significantly low under other baselines. It may be that FedGen's model is designed based on the mnist dataset, and lacks scalability to the relatively large datasets such as CIFAR-10, CIFAR-100 and Tiny-ImageNet, so the performance of FedGen is poor. Additionally, we can see that our extended and unified computation-aware framework, with the increase of computation-unconstrained clients ratio \(\lambda\), model performance continues to improve. We will discuss the upper limits of this improvement in our extended and unified computation-aware framework in section \ref{unfied}. We also experiment in various pathological heterogeneity settings of CIFAR-10 and CIFAR-100. Table \ref{tab:result2} shows the result.


The above experiments show that our method can improve the performance of the global model in the non-iid setting under constrained computation through knowledge filling and knowledge sampling. Meanwhile, to deal with more complex real-world situations, we also propose an extended and unified computation-aware framework, and the performance of our method can be further improved with an increasing computation-unconstrained clients ratio \(\lambda\).

\subsection{Extended and Unified Framework}\label{unfied}
In this section, we analyze the impact of the proportion of computation-unconstrained clients ($\lambda$) on the performance of our extended and unified computation-aware framework. As shown in Table~\ref{tab:result2}, increasing $\lambda$ generally improves model accuracy across pathological heterogeneity, scalability, and robustness settings. However, the improvement exhibits diminishing returns, with performance approaching an upper bound when $\lambda$ reaches 25\%–40\%.
\begin{table*}[ht]
\caption{Mean accuracy (\%) under pathological heterogeneity (shards), scalability (number of clients), and robustness (noise injection) settings across CIFAR-10, CIFAR-100, and Tiny-ImageNet. Bold denotes the best result. Values are reported as mean (standard deviation).}
\label{tab:result2}
\centering
\begin{tabular*}{\linewidth}{@{\extracolsep{\fill}}lccccccccccccc@{}}
\toprule
& \multicolumn{4}{c}{Pathological Heterogeneity} & \multicolumn{2}{c}{Scalability} & \multicolumn{3}{c}{Robustness} \\
\cmidrule(lr){2-5} \cmidrule(lr){6-7} \cmidrule(lr){8-10}
& \multicolumn{2}{c}{CIFAR-10} & \multicolumn{2}{c}{CIFAR-100} & \multicolumn{2}{c}{CIFAR-10} & \multicolumn{3}{c}{Tiny-ImageNet} \\
\cmidrule(lr){2-3} \cmidrule(lr){4-5} \cmidrule(lr){6-7} \cmidrule(lr){8-10}
Method & Shards (4) & Shards (8) & Shards (6) & Shards (12) & 50 clients & 100 clients & Noise (0.01) & Noise (0.05) & Noise (0.1) \\
\midrule
FedAvg~\cite{DBLP:conf/aistats/McMahanMRHA17} & 63.77(1.24) & 78.46(0.91) & 14.91(0.53) & 31.74(0.78) & 68.92(1.04) & 60.63(3.16) & 31.58(0.67) & 29.12(0.79) & 27.21(0.86) \\ 
FedProx~\cite{DBLP:conf/mlsys/LiSZSTS20} & 64.10(2.67) & 77.98(0.53) & 13.64(0.48) & 30.48(0.80) & 68.63(1.19) & 60.53(3.17) & 32.46(0.87) & 30.93(0.59) & 29.10(0.40) \\ 
Scaffold~\cite{pmlr-v119-karimireddy20a} & 76.75(0.64) & 79.86(0.79) & 12.68(0.66) & 26.94(0.13) & 72.39(0.91) & 62.30(1.04) & 34.18(1.51) & 32.88(1.69) & 31.14(0.98) \\ 
MOON~\cite{li2021model} & 78.98(0.47) & 80.42(0.73) & 18.46(0.34) & 37.46(0.96) & 69.24(1.07) & 60.53(3.17) & 33.78(2.51) & 31.39(1.86) & 30.55(1.78) \\ 
FedGen~\cite{pmlr-v139-zhu21b} & 67.14(1.03) & 76.58(1.24) & 11.39(0.87) & 29.98(0.98) & 66.84(2.08) & 60.78(1.35) & 30.48(1.86) & 29.17(1.15) & 26.92(1.31) \\ 
FedDyn~\cite{acar2021federatedlearningbaseddynamic} & 73.29(1.68) & 77.38(1.45) & 14.29(0.63) & 34.98(1.06) & 63.16(1.37) & 63.78(1.43) & 35.48(1.19) & 32.17(1.21) & 29.94(1.07) \\ 
\midrule
Ours & \textbf{80.74}(0.66) & \textbf{82.62}(0.30) & \textbf{20.47}(0.46) & \textbf{38.54}(0.26) & \textbf{75.27}(0.56) & \textbf{66.28}(0.34) & \textbf{39.95}(0.33) & \textbf{39.67}(0.46) & \textbf{38.89}(0.26) \\
\bottomrule
\end{tabular*}
\end{table*}


\begin{table*}[h]
\caption{Ablation Study.}
\label{tab:result3}
\centering
\begin{tabular*}{\linewidth}{@{\extracolsep{\fill}}lcccccccccc@{}}
\toprule
Method & \multicolumn{2}{c}{Knowledge Sampling} & \multicolumn{2}{c}{Knowledge Filling} & \multicolumn{4}{c}{Knowledge Replay} & \multicolumn{2}{c}{Dataset} \\
\cmidrule(lr){2-3} \cmidrule(lr){4-5} \cmidrule(lr){6-9} \cmidrule(lr){10-11}
& Random & Knowledge & Alignment & Drop & 0\% & 10\% & 70\% & 90\% & CIFAR-10 & CIFAR-100 \\
\midrule
M1 & \checkmark & & & & \checkmark & & & & 75.16(0.29) & 34.13(0.17) \\
M2 & & \checkmark & & & \checkmark & & & & 76.84(0.38) & 36.26(0.30) \\
M3 & & \checkmark & \checkmark & & \checkmark & & & & 74.51(0.34) & 34.13(0.31) \\
M4 & & \checkmark & \checkmark & \checkmark & \checkmark & & & & 80.84(0.26) & 47.82(0.24) \\
M5 & & \checkmark & \checkmark & \checkmark & & & & \checkmark & 79.84(0.52) & 46.89(0.33) \\
M6 & & \checkmark & \checkmark & \checkmark & & & \checkmark & & 80.14(0.46) & 47.29(0.48) \\
M7 (Ours) & & \checkmark & \checkmark & \checkmark & & \checkmark & & & \textbf{81.62}(0.30) & \textbf{48.65}(0.20) \\
\bottomrule
\end{tabular*}
\end{table*}

\subsection{Scalability and Robustness}
To show the scalability of our method, we conduct two experiments with 50 and 100 clients on the the setting of CIFAR-10 DIR(0.5). Table ~\ref{tab:result2} shows that in 50 and 100 clients setting, our method still performs best, which indicates that our method has scalability.
On the other hand, we simulate the noise and absence that may occur in the real image by controlling the proportion of gaussian noise added to the real image of Tiny-ImageNet. We find that most algorithms show model degradation under noise setting, which may be caused by the coupling of data enhancement and noise information during model training. At the same time, we find that there is almost no model degradation in our method, indicating that our method is more robust.

\subsection{Ablation Study}
In this section, we delve into the role of the different components in our approach. Table ~\ref{tab:result3} shows the results of our approach on the CIFAR-10 and CIFAR-100 datasets with different components. M1, M2, M3, M4, M5, M6 and M7 represent our approach \textbf{FBL} when different components are adopted. M7 is adopted by our method.

\subsubsection{Analysis of Knowledge sampling.}
Due to the constrained computation premise, i.e., the number of client samples remains fixed, our method must sample on the data-excessive classes.
Therefore, we compare the performance of our method on CIFAR-10 and CIFAR-100 using random sampling and knowledge sampling methods respectively. Table ~\ref{tab:result3} indicates that the knowledge sampling based on loss method is better than the random sampling method on CIFAR-10 and CIFAR-100. This shows that knowledge sampling based on loss is meaningful. The comparison between M1 and M2 confirms our conclusion.

\subsubsection{Analysis of knowledge Filling.}
\begin{figure}[h]
\centering
\includegraphics[width=\linewidth]{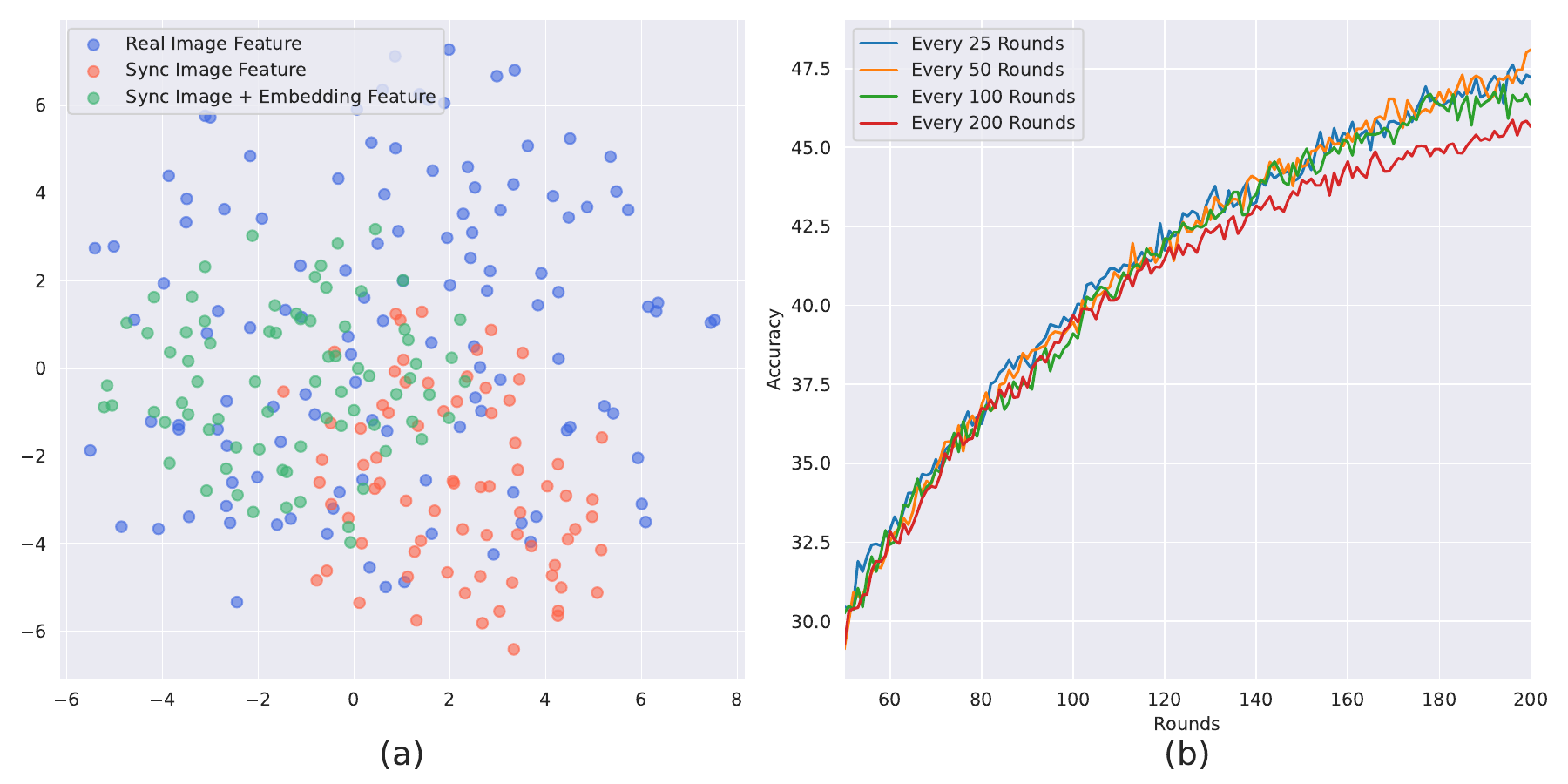}
\caption{(a) Visualization of Knowledge Alignment Strategy and Knowledge Drop Strategy. It brings the synthetic data's domain closer to the domain of the real data.
 (b) Visualize of the results of our method using different cycles.}
\label{fig:round}
\end{figure}

In fact, we believe that there is a domain gap between the synthetic data and the real data, so the knowledge alignment strategy of embedding is used to bridge this gap. Subsequently, we introduced the knowledge drop strategy to regularize the learning of embeddings. As shown in Table ~\ref{tab:result3}, M3 and M2 indicate that without knowledge drop regularization, the embeddings can become a negative optimization due to the lack of constraints. When knowledge drop regularization is applied, as demonstrated by M4, the performance significantly improves compared to M3 and M2. This highlights the effectiveness of the two strategies we proposed. Figure~\ref{fig:round} visualizes the domain gap between synthetic and real data that is eased by the regularized embeddings. The two strategies bring the synthetic data's domain closer to the domain of the real data. Additionally, to facilitate the transformation of synthetic data to real data, we do not use data augmentation for the synthetic data but apply it to the real data. This approach ensures the model's performance and accelerates convergence.

\noindent{\textbf{Analysis of periodic knowledge replay.}}
In order to reduce the loss of knowledge diversity caused by data downsampling, our method adopts the strategy of periodic knowledge replay. Periodic knowledge replay has two parameters, one is the periodic parameter which is every \(\ m \) round, and the other is the ratio of the knowledge replay which is proportion of samples retained from the previous cycle's dataset \(\gamma \). Through the ablation analysis experiments of different cycle \(\ m \) rounds from Figure~\ref{fig:round}, we confirm the validity of periodic knowledge replay. As can be seen from table~\ref{tab:result3}, the ratio of knowledge replaying is a trade-off parameter, which can ensure a certain diversity of data with the addition of new data, and at the same time prevent catastrophic forgetting of the model through the replaying of old data. We can find that in our experiment, Replay(10\%) has the best effect, and this parameter is neither the highest nor the lowest among all parameters. The above experiments also show that this is a trade-off parameter. The above two sets of experiments demonstrate the effectiveness of our periodic knowledge replay component.

\begin{table}[t]
\caption{The effect of different proportions of Computation-unconstrained clients on our method under CIFAR-10 and CIFAR-100. \textbf{Bold} fonts highlight the best accuracy.}
\label{tab:result4}
\centering
\begin{tabular*}{\linewidth}{@{\extracolsep{\fill}}lccccc@{}}
\toprule
Dataset & \multicolumn{2}{c}{CIFAR-10 (\%)} & \multicolumn{2}{c}{CIFAR-100 (\%)} \\
\cmidrule(lr){2-3} \cmidrule(lr){4-5}
$\alpha$ & 0.1 & 0.5 & 0.1 & 0.5 \\
\midrule
Ours            & 76.06 (0.24) & 84.58 (0.08) & 36.52 (0.30) & 52.29 (0.16) \\
Ours (10\%)     & 78.62 (0.33) & 84.71 (0.28) & 43.78 (0.44) & 54.61 (0.25) \\
Ours (25\%)     & 79.21 (0.15) & 84.95 (0.28) & 48.59 (0.37) & 57.61 (0.45) \\
Ours (40\%)     & 79.69 (0.28) & 85.38 (0.32) & 48.23 (0.11) & \textbf{57.91} (0.31) \\
Ours (50\%)     & \textbf{79.88} (0.24) & 85.89 (0.21) & 48.47 (0.26) & 57.36 (0.56) \\
Ours (75\%)     & 79.81 (0.34) & \textbf{85.92} (0.17) & \textbf{48.96} (0.39) & 57.45 (0.68) \\
\bottomrule
\end{tabular*}
\end{table}

\noindent{\textbf{Analysis of extreme edge-side devices.}}
Consider an extremely constrained federated learning scenario where some edge-side devices are unable to run even the Stable Diffusion models that can be executed on a Raspberry Pi. For these extreme clients, we refrain from using the generative model and maintain consistency with FedAvg. In this resource-limited context, we conducted experiments on the CIFAR-10 dataset, setting the Dirichlet parameter \(\alpha\) to 0.3. We designated 30\% of the clients as extremely constrained, meaning they were incapable of generating images. The final result achieved was 80.8 (0.57), which is slightly lower than the 81.62 (0.3) obtained when all clients were capable of image generation. This demonstrates the scalability and robustness of our method.

\begin{figure}[h]
\centering
\includegraphics[width=0.8\linewidth]{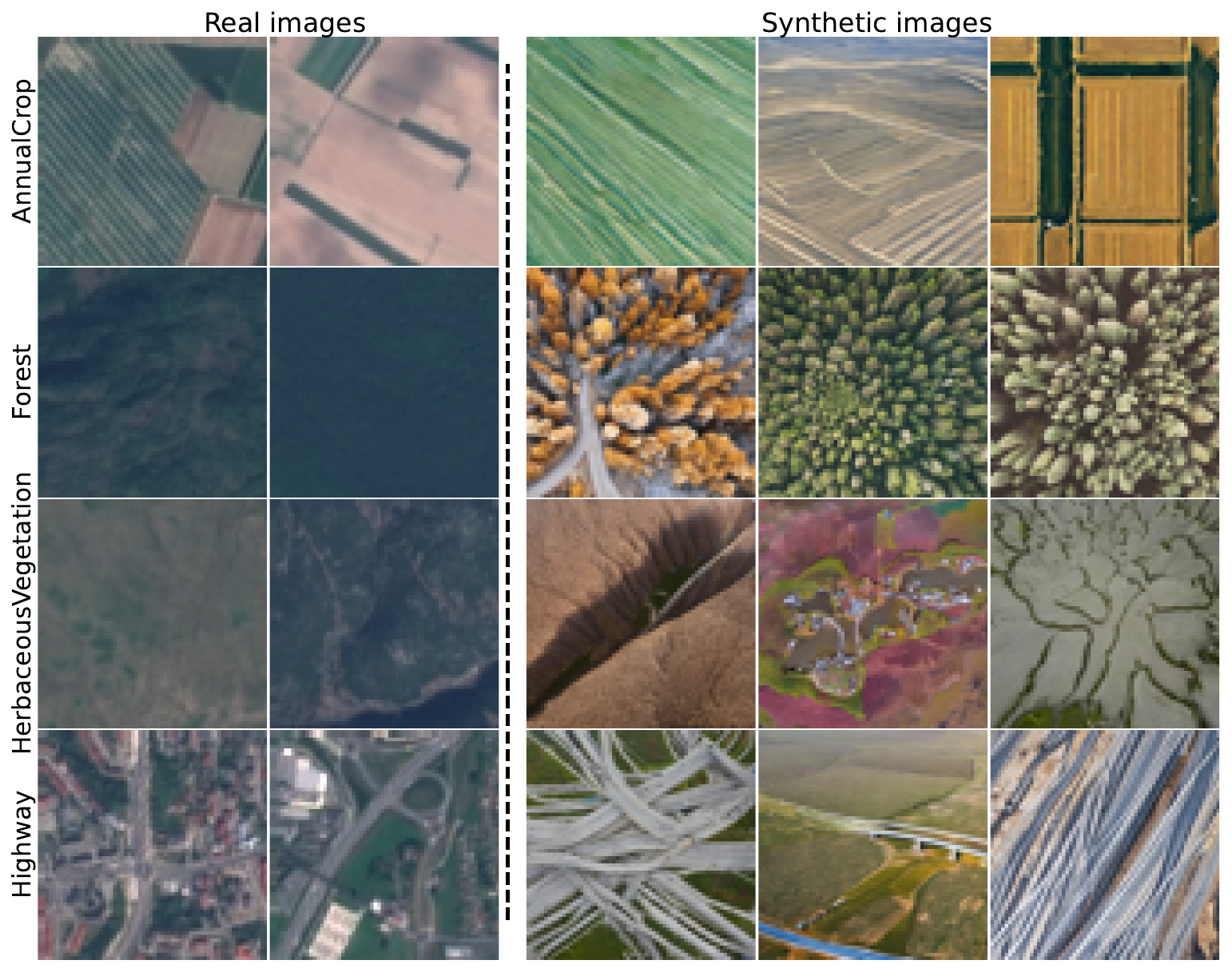}
\caption{Real and synthetic images on the EuroSAT dataset.}
\label{fig:EuroSAT}
\end{figure}

\noindent{\textbf{Results in more challenging domains.}}
The EuroSAT~\cite{helber2019eurosatnoveldatasetdeep} dataset presents unique challenges due to its high variability in image content and the complexity of satellite imagery. These factors make it an ideal testbed for assessing the robustness and adaptability of our method. To further validate the paradigm extensibility of our approach, we conducted experiments on the EuroSAT satellite imagery dataset. By setting the Dirichlet parameter \(\alpha\) to 0.3, our method achieved a result of 84.42 (0.52), slightly surpassing FedAvg's performance of 82.7 (1.51). This slight improvement over FedAvg demonstrates that our approach can maintain a competitive edge even under more demanding conditions. Figure.~\ref{fig:EuroSAT} illustrates the visualization results of both real and synthetic images on the EuroSAT dataset.

\section{Limitations and Future Prospects}

\noindent{\textbf{Limitations.}}
In the FBL paradigm, the effectiveness of data-scarce and data-missing Knowledge Filling relies on the quality of data generated by generative models. However, in more challenging domains, the data generated by these models is often inferior compared to that in more common domains. This discrepancy results in the advantages of FBL being less pronounced in more challenging domains.

\noindent{\textbf{Future prospects.}}
Several studies~\cite{ali2022spotfakelungsgenerating,li2024lightnightmulticonditiondiffusion} have attempted to apply generative models to address data issues in difficult domains, achieving promising results. We believe that with the ongoing development of generative models and scaling laws, the capabilities of these models will continue to improve, potentially bridging this gap. On the other hand, the increasing computational power of edge devices, such as the deployment of MLLMs~\cite{wen2025enhanced} and generative models on mobile devices, like SnapFusion and MobileDiffusion, demonstrates the potential of FBL.

\section{Conclusion}\label{conclusion}
In this paper, we propose \textbf{FBL} to achieve sample balance by knowledge sampling and knowledge filling. In the knowledge filling component, we propose a Knowledge Alignment Strategy and a Knowledge Drop Strategy. Finally, we extended our framework to adapt to the real and complex scenarios. 


\section*{Acknowledgement.}

This work makes use of the CIFAR-10 and CIFAR-100 datasets (https://www.cs.toronto.edu/~kriz/cifar.html
). The authors of this work confirm that the use of the above datasets in this work is strictly limited to academic research purposes and does not involve any commercial activities.

\bibliographystyle{IEEEtran}
\bibliography{ref}

%

\section{Biography Section}

\end{document}